# Dynamic technology impact analysis: A multi-task learning approach to patent citation prediction

*Youngjin Seol[a], Jaewoong Choi[b], Seunghyun Lee[a], Janghyeok Yoon[a,\*]*

[a]Department of Industrial Engineering, Konkuk University, 120 Neungdong-ro, Gwangjin-gu, Seoul 05029, Republic of Korea

[b]Computational Science Research Center, Korea Institute of Science and Technology, 5 Hwarang-ro 14-gil, Seongbuk-gu, Seoul 02792, Republic of Korea

[\*]Correspondence to Janghyeok Yoon (janghyoon@konkuk.ac.kr)

**Abstract:**

Machine learning (ML) models are valuable tools for analyzing the impact of technology using patent citation information. However, existing ML-based methods often struggle to account for the dynamic nature of the technology impact over time and the interdependencies of these impacts across different periods. This study proposes a multi-task learning (MTL) approach to enhance the prediction of technology impact across various time frames by leveraging knowledge sharing and simultaneously monitoring the evolution of technology impact. First, we quantify the technology impacts and identify patterns through citation analysis over distinct time periods. Next, we develop MTL models to predict citation counts using multiple patent indicators over time. Finally, we examine the changes in key input indicators and their patterns over different periods using the SHapley Additive exPlanation method. We also offer guidelines for validating and interpreting the results by employing statistical methods and natural language processing techniques. A case study on battery technologies demonstrates that our approach not only deepens the understanding of technology impact, but also improves prediction accuracy, yielding valuable insights for both academia and industry.

# 1. Introduction

Technology impact analysis—defined as the impact of technology on society, particularly unintended, indirect, or delayed consequences, during its introduction, development, or modification— is essential for technology management, enabling proactive identification and mitigation of potential risks (Ayres, 1969; Group, 2004; Lee et al., 2012). Earlier, expert-centric approaches, such as the Delphi method and scenario modeling, have been predominantly employed to generate forward-looking insights into the impact of technology (Lee, 2021). As the proliferation of technologies accelerates and innovation cycles shorten, these qualitative methods have been replaced by data-driven methods. In this context, patents have been regarded as valuable data sources because of their technical content, high quality, reliability, and accessibility (Ernst, 1997; Lee et al., 2020). Patent citation information is primarily used for technology impact analysis as it positively correlates with technological value (Arts and Veugelers, 2015; Harhoff et al., 1999; Lee et al., 2012; Trajtenberg, 1990b).

Patent citation-based methods for technology impact analysis can be summarized as probabilistic and machine learning (ML)-based approaches. First, the probabilistic approach to technology impact analysis often uses curve-fitting techniques and stochastic models to project future trends by estimating the patent citation counts as quantitative indicators (Lee et al., 2012; Shin et al., 2013). However, because these methods require patents to accumulate citations over time, they typically highlight current key technologies rather than early stage innovations (Lee et al., 2018). In response, some researchers have developed ML-based approaches that utilize early patent information to predict citation counts, demonstrating the effective prediction of a patent's technological impact (Chung and Sohn, 2020; Hong et al., 2022; Lee et al., 2018). Nevertheless, considering that technology impact evolves over time and that there are interdependencies between impacts across different periods, where earlier impacts influence later ones and vice versa, the structure and design of existing ML-based approaches need to be improved (Ernst, 1997; Lee et al., 2011; Lee et al., 2017).

As a remedy, we propose a multi-task learning (MTL) approach to enhance the understanding of the time-variant nature of technology impact and the interrelationships between impacts across different time periods. MTL is an ML approach in which multiple related tasks are learned simultaneously, sharing knowledge between tasks to improve overall performance. By applying MTL, we enhance the prediction performance of technology impact across different time periods through knowledge sharing while also tracking the evolution of technology impact over time. In particular, we address the following research questions: (1) Does the technology impact vary over time? (2) Does the impact of different time periods influence each other? (3) Why does the impact of technology change over time? At the core of our approach are (i) the quantitative definition and pattern identification of technology impact through citation analysis across different time periods; (ii) the design and development of an MTL model to predict citation counts from multiple patent indicators over time; and (iii) the interpretation of key input indicator shifts and patterns across time periods using the SHapley Additive exPlanation (SHAP) method.

We applied the proposed approach to 10,851 patents in the field of battery technology. The case study demonstrated that the proposed approach is effective in understanding the dynamic nature of technology impact, as well as the interrelationships between technological impacts over different time periods. In addition, the use of SHAP analysis revealed that scientific knowledge of technologies is the most important feature for long- and short-term predictions, and has positive effects on breakthrough technology. To statistically validate the predictions, we compared the post-hoc patent value indicators of high- and lower-impact technologies, revealing significant differences in patent maintenance periods, technology transfer counts, and patent family size. We provided a practical implementation guideline combining natural language processing with our approach to present technology impact dynamics more clearly, identifying 22 battery sub-domains and revealing high-impact technologies, such as 'direct methanol fuel cells' and 'energy storage devices.' The proposed approach is expected to have academic implications as an initial attempt to understand the nature of the technology impact with MTL methods, and the systematic process and quantitative outcomes are anticipated to assist intellectual property management and R&D planning for practitioners.

The remainder of this paper is organized as follows. Section 2 outlines the background of this study. Section 3 describes the proposed methodology, which is illustrated using the case study in Section 4. Section 5 presents the guidelines for validating and implementing the proposed approach. Finally, Section 6 addresses the limitations of this study and offers directions for future research.

## 2. Background

The impact of technology is generally assessed by the extent to which it serves as prior art for subsequent technological development (Arts and Veugelers, 2015; Harhoff et al., 1999; Trajtenberg, 1990b). Technologies with a strong impact are considered breakthroughs in value creation and growth (Arts and Veugelers, 2015; Briggs, 2015; Scherer and Harhoff, 2000; Schumpeter, 1942). The strategic importance of forecasting technological changes has become evident (Lee et al., 2012), and the key to technology forecasting is identifying the current technologies that will drive technological changes in the coming years (Porter, 1991; Watts and Porter, 1997). However, predicting the future impact of technology is challenging because of the complexity and uncertainty of the technological environment (Ayres, 1969). In this regard, patent data could provide quantitative methods for forecasting technology impact, as they contain credible and detailed information on technologies derived from a large number of patents (Lee, 2021).

Early attempts predicted the future impact of technologies by analyzing historical citation data and projecting future trends based on the estimated citation counts. Considering that technology impacts are unstable and dynamic, changing over time throughout their lifetime, unique to each technology (Lee et al., 2012), some researchers have adopted stochastic processes (Jang et al., 2017; Lee et al., 2012; Lee et al., 2016) and curve-fitting methods (Shin et al., 2013). A tenet of these studies is that shifts in patent information can track a technology's progression through its life cycle stages, and the relationship between the technology impact at different stages allows for predicting the individual technology impact of an invention. Despite the valuable contributions of previous studies in providing quantitative methods for predicting technology impacts, reliance on the accumulation of citation data limits their applicability when relevant inventions are in the early stages of technology development, without sufficient historical citation data (Lee et al., 2018).

As a remedy, ML-based methods have been suggested as predictive methods for technology impact analyses. In a pioneering study, Lee et al. (2018) the use of ML models to model the nonlinear relationships between quantitative patent indicators and patent citation counts, thereby allowing the identification of emerging technologies at an early stage. Similarly, Chung and Sohn (2020) suggested a multi-modal model using both textual information and quantitative patent indicators as inputs for predicting patent citation counts. Recently, Hong et al. (2022) enhanced the understanding of textual information for technology impact analysis by adopting convolutional neural networks for unstructured patent texts. These studies, despite making significant contributions to effectively expressing the technical characteristics of patents and modeling their technological impact, have overlooked certain aspects of the nature of technology impact, specifically, the time-variant nature of technological impact and the interrelationships between technological impacts over different time periods. Of course, some studies have taken multiple observation points for technology impact, but they have modeled these independently, failing to fully capture the characteristics of technology impact. Table 1 presents a comparison between previous studies and the current study.

**Table 1.** Comparison between prior studies and the current study.

|  | **Focus of analysis** | **Concept** | **Result** |
|---|---|---|---|
| Probabilistic approach(Lee et al., 2012; Lee et al., 2016; Shin et al., 2013) | Providing insights into future technology impacts by integrating patent citation analysis with a stochastic process. | A stochastic process and curve fitting-based approach to capture patent citation pattern | Static snapshot of current key technologies |
| ML-based approach (Chung and Sohn, 2020; Hong et al., 2022; Lee et al., 2018) | Identifying future technology impacts using early stages patent indicators | A ML-based approach to identify future technology impacts by capturing the early-stage patent information | Static snapshot of breakthrough technologies |
| Current study | Providing an enhanced understanding of the dynamics of technology impact and the interrelationships between technological impacts over different time periods | Applying MTL to predict future technology impact across different time horizon by sharing the knowledge within the same machine | Dynamic snapshot of breakthrough technologies |

# 3. Methodology

The overall process of the proposed approach involves four discrete steps (Figure 1): (1) data collection and preprocessing, (2) definition and extraction of patent indicators, (3) estimation of dynamic technology impact, and (4) model validation and result interpretation.

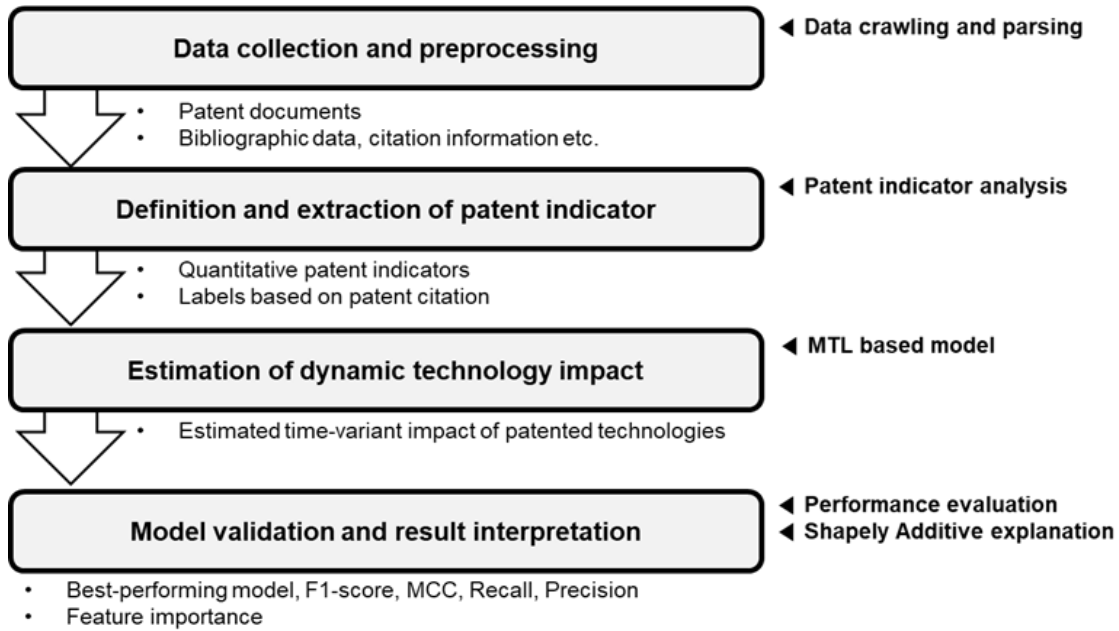

**Figure 1**. Overall process of the proposed approach.

## 3.1. Data collection and preprocessing

Once the target domain is defined, relevant patents can be collected from databases such as the United States Patent and Trademark Office (USPTO) or the Derwent Innovations Index. Patents contain both structured data, such as metadata, and unstructured data, such as abstracts and claim texts, and are typically provided in semi-structured formats, such as HTML or XML. These are parsed according to the data type and stored in the relational database. Key information in this database can be categorized into three main types: bibliographic information, textual information, and citation data. Bibliographic information includes patent registration numbers, filing/registration dates, and classification codes, which are used to systematically define the analysis periods and scopes. Textual information consists of titles, abstracts, and claims, which provide detailed descriptions of the technology and its scope of protection. Finally, citation data represent the technological dependencies between patents, and are primarily stored as an edge list between citing and cited patents.

**3.2. Definition and extraction of patent indicators**

This study employs multiple patent indicators that represent various technological characteristics and can be defined at the time of patent registration. Through an extensive literature review, we identified 44 quantitative indicators that are directly or potentially associated with the technology impact (Table 2). These indicators can be grouped into six categories, each offering a different perspective: (1) scope and coverage, (2) priority, (3) development effort and capabilities, (4) completeness, (5) technology environment, and (6) prior knowledge. Of the total, 28 are quantitative indicators and the remaining 16 are nominal variables measured for each technology field. Details of each indicator are described in Appendix A1.

**Table 2.** Summary of quantitative patent indicators used as input variables.

| Category | Indicator | Operational description | Index |
|---|---|---|---|
| Scope and coverage (Fischer and Leidinger, | International scope of prior patents | Number of different nations in the backward citation information of the patent | SC_1 |

| | Scope of core technological components | Number of independent claims in the patent | SC_2 |
|---|---|---|---|
| 2014; Jeong et al., 2016; Lanjouw and Schankerman, 2001, 2004; Lee et al., 2009) | Scope of peripheral technological components | Number of dependent claims in the patent | SC_3 |
| | Average specificity of core technological components | Average numbers of words contained in independent claims in the patent | SC_4 |
| | Technical coverage in the system | Number of different IPCs assigned to the patent | SC_5 |
| Priority (Choi et al., 2020; Su et al., 2011) | Priority intensity | Number of priority patents of the patent | PR_1 |
| | International priority range | Number of different nations where the patent has its priorities | PR_2 |
| Development effort and capabilities (Ernst, 2003; Lai and Che, 2009; Ma and Lee, 2008) | Size of contributors | Number of applicants involved with the patent | DEC_1 |
| | Contributions of foreign applicants | Number of non-US applicants involved with the patent | DEC_2 |
| | International distribution of contributors | Number of different nations of applicants involved with the patent | DEC_3 |
| | Efforts of inventors | Number of inventors involved with the patent | DEC_4 |
| | Efforts of foreign inventors | Number of non-US inventors involved with the patent | DEC_5 |
| | International cooperation degree | Number of different nations of inventors involved with the patent | DEC_6 |
| Completeness (Hikkerova et al., 2014) | Grant time lag | Patent examination period | CP_1 |
| | Specificity of technical summary | Number of words contained in the patent's abstract | CP_2 |
| Technology Environment (Bierly and Chakrabarti, 1996; Choi et al., 2023; Fabry et al., 2006; Kayal and Waters, 1999; Lai and Che, 2009; Trappey et al., 2012) | Activity of technology fields | Average number of patents issued yearly in the relevant IPCs | TE_1 |
| | Size of technology fields | Average number of cumulative patents issued in the relevant IPCs | TE_2 |
| | Competitiveness of technology fields | Average number of applicants issuing patents by issuance year in the relevant IPCs | TE_3 |
| | Technology fields in the system | Frequency of sections A to H of the patent | TE_4 |
| | Growth speed | Median gap of the filing date between prior patents and the patent, as known as technology cycle time | TE_5 |
| | Scientific knowledge | Number of non-patent citations | PK_1 |

| | Technological recombination | The ratio of the IPC of the patent to the IPC of prior patents | PK_2 |
|---|---|---|---|
| | Technology fields of prior patents in the system | Frequency of sections A to H of prior patents | PK_3 |
| Prior knowledge (Chung et al., 2021; Cozzens et al., 2010; Harhoff et al., 1999; Kwon and Geum, 2020; Lee et al., 2018; Lee and Lee, 2019; Meyer, 2006; Rotolo et al., 2015; Seol et al., 2023; Trajtenberg, 1990a) | Technological breadth | Breadth of technology fields on which the patent is based | PK_4 |
| | Dependency on homogeneous technologies | Number of backward citations in the battery technology field | PK_5 |
| | Prior knowledge of inventors | Average number of patents issued by inventors | PK_6 |
| | Prior experience of assignees | Average number of patents issued by assignees | PK_7 |
| | Core area know-how | Number of patents in a technology field of interest issued by assignees | PK_8 |
| | Peripheral area knowhow | Number of patents in other technology fields issued by assignees | PK_9 |
| | Prior knowledge | Number of backward citations | PK_10 |

## 3.3. Estimation of dynamic technology impact

In this step, MTL is employed to effectively handle the task of predicting time-variant technology impacts. By concurrently learning tasks with varying time horizons within a shared network, the model enhances generalization by capturing both task-specific and shared knowledge across tasks (Caruana, 1997). In our case, the MTL model can leverage the correlations between different time horizon technology impacts, reflecting the distinctiveness of each task. The MTL model can be designed as an input, shared, and task-specific layer (Figure 2). In the input layer, the model receives features that represent the technological characteristics of the individual patents. The shared layers capture the shared knowledge across tasks related to technology impact prediction over multiple time horizons. Finally, the task-specific layers learn knowledge tailored to each task and optimize the predictions for different time horizons.

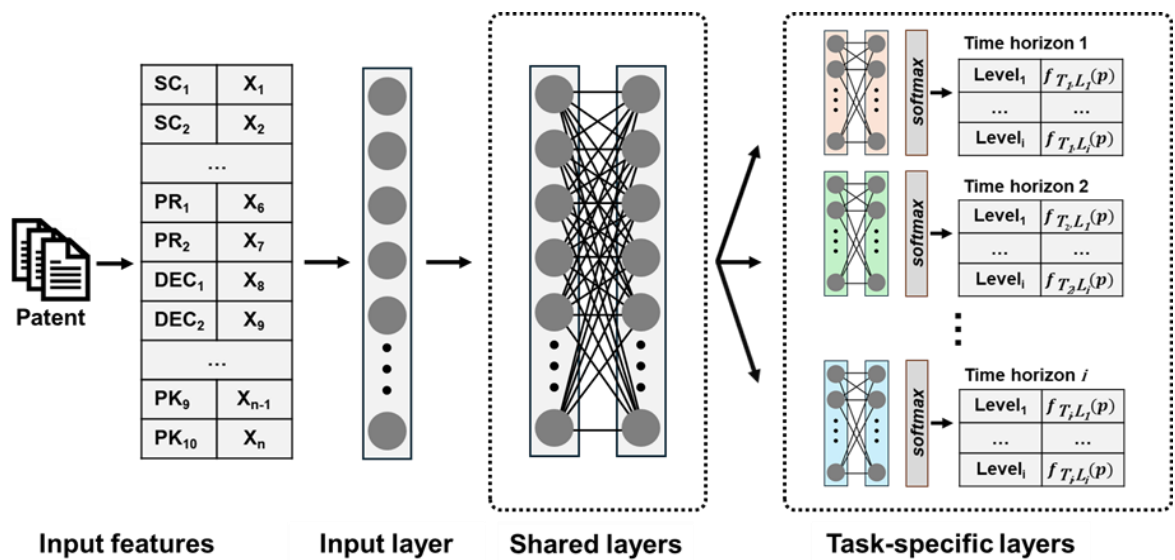

**Figure 2.** Graphical illustration of the proposed MTL model.

The model predicts the technology impact level for each time horizon T∈ {T1, T2, T3, …}, given the input patent. For each time horizon $T$, the model passes through the task-specific layer and outputs a logit $z_{T,i}$ which represents the logit value for influence level $i$ at time horizon T. These logits are then passed through a SoftMax function to convert them into probabilities for each technology impact level. Finally, the technology impact of each time horizon is determined by applying argmax to the probabilities and selecting the impact level with the highest probability (Equations 1-2). $f_{t,i}(p)$ represents the probability that patent p belongs to technology impact level i at time horizon t;

$$f_{T,i}(p) = \frac{\exp(z_{T,j})}{\sum_j \exp(z_{T,j})} \tag{1}$$

$$Impact\ Level_T = \arg\max_i f_{T,i}(p) \tag{2}$$

The shared and task-specific parameters are refined and optimized using backpropagation based on the loss function for each individual task (Equation 3). The model is trained to minimize the weighted loss for each task. $\theta_{sh}$ and $\theta_{T_i}$ represents the shared parameters and the task-specific parameters, respectively. $L_{T_i}$, $D_{T_i}$ and $w_i$ are the loss for each task, the dataset for each task, and the weight for each task's loss. The model learns parameters that minimize the weighted sum of losses across all the datasets.

$$\min_{\theta_{sh}, \theta_{T_1} \dots \theta_{T_n}} \sum_{i=1}^{n} w_{T_i} * L_{T_i}(\{\theta_{sh}, \theta_{T_i}\}, D_{T_i}) \tag{3}$$

### 3.4 Model validation and result interpretation

In this step, a comprehensive performance evaluation is conducted to identify the best-performing model using traditional performance metrics, such as precision, recall, and F1-score. In our case, recall indicates how well the model identifies all patents of a grade, which is crucial for detecting important but less frequent grades, while precision shows the accuracy of the model's predictions for each grade, which is important for reducing false positives. The F1-score balances precision and recall, offering a single metric that accounts for both false positives and false negatives, and provides a fuller picture of the model's performance (Appendix A2). In addition, we employed two measures, the diagnostic odds ratio (DOR) and Matthews correlation coefficient (MCC), to address both class imbalance and overall classification quality. DOR evaluates the odds of a correct classification versus an incorrect one, providing insight into the model's ability to discriminate between classes, whereas MCC accounts for true and false positives and negatives, offering a balanced measure even in imbalanced datasets (Equations 4-5).

$$DOR_i = \frac{TP_i \times TN_i}{FP_i \times FN_i} \quad (4)$$

$$MCC_i = \frac{TP_i \times TN_i - FP_i \times FN_i}{\sqrt{(TP_i+FP_i)(TP_i+FN_i)(TN_i+FP_i)(TN_i+FN_i)}} \quad (5)$$

Following the performance evaluation, SHAP is employed to interpret the model results by analyzing patterns of technology impact shifts over time. SHAP enables a deeper analysis at the patent level, allowing patents to be grouped based on changing impact trends or time horizons. This facilitates the identification of key characteristics of technologies whose impacts increase or decrease over time. Rooted in cooperative game theory, SHAP attributes feature importance fairly, making it highly effective in uncovering common patterns across patent groups (Scott and Su-In, 2017). Its model-agnostic nature, along with specialized implementations, such as tree SHAP, kernel SHAP, and deep SHAP, makes it applicable to a wide range of ML models. By applying SHAP to our models, it would be possible to compare pivotal input features for short- and long-term technology impacts, thereby gaining valuable insights into the dynamics of technology impacts.

## 4. Empirical analysis and results

We conducted a case study of battery technology for three primary reasons. First, considering that battery technologies drive key industries, such as electric vehicles and renewable energy, technology forecasting helps prioritize investments and maintain competitiveness in these fast-growing sectors (Larcher and Tarascon, 2015). Second, as batteries are essential for clean energy transitions, identifying emerging technologies enables policymakers to support innovations that align with sustainability goals through funding and regulation (Kim et al., 2023). Finally, industrial stakeholders need scientifically grounded forecasts of battery technology advancements to optimize resource allocation, supply chains, and workforce planning, ensuring readiness for technological shifts. Failures in R&D decision-making can result in resource misallocation and delayed green innovation, hindering progress toward sustainability goals. A total of 10,851 patents classified under IPC code H01M and registered in the USPTO between 2006 and 2012 were collected.

### 4.1. RQ1: Does the technology impact vary over time?

To answer RQ1— whether the impact of a technology changes over time, indicating the presence of dynamic behavior— we tracked the number of citations received by patents over three time horizons: three, five, and ten years following their issuance. We found that the distribution of patents based on these forward citation counts was markedly right-skewed (Table A. 1); the majority of patents received few or no citations, while a small subset of patents received a disproportionately high number of citations. We defined the technology impact class of patents based on the number of forward citations for each time horizon (Table 3). According to the stanine rating system, patents classified as first grade, representing approximately the top 3.6–4%, were labeled as breakthrough technologies. Patents in the second and third grades were categorized as valuable technologies, whereas the remaining patents were classified as moderate technologies.

**Table 3.** Class of technology impact based on patent citation counts.

| Class | Short-term | Mid-term | Long-term | Alias |
|---|---|---|---|---|
| Breakthrough technologies | Equal to or exceeding 4 (3.67%) | Equal to or exceeding 9 (3.61%) | Equal to or exceeding 24 (4.00%) | BT |
| Valuable technologies | 2-3(7.82%) | 3-8(14.03%) | 6-23(18.10%) | VT |
| Moderate technologies | 0-1(88.51%) | 0-2(82.36%) | 0-5(77.89%) | MT |

Examining the temporal patterns of changes in technology impact ratings reveals the presence of intriguing patent groups, as illustrated in Figure 3 and Appendix A3. For instance, certain patents consistently receive citations over time, indicating their status as breakthrough technologies, while others show a gradual decline or increase in impact over time (Lee et al., 2017). We marked sustained-impact patents that continuously exerted significant influence and remained highly relevant over extended periods with blue lines. Peak-and-fade patents, which initially have a significant impact but subsequently diminish in relevance and influence, are represented by the 'yellow line.' There are late-blooming patents that initially exhibit minimal impact but gain recognition as valuable assets once their potential is validated in the market ('green line' in Figure 3).

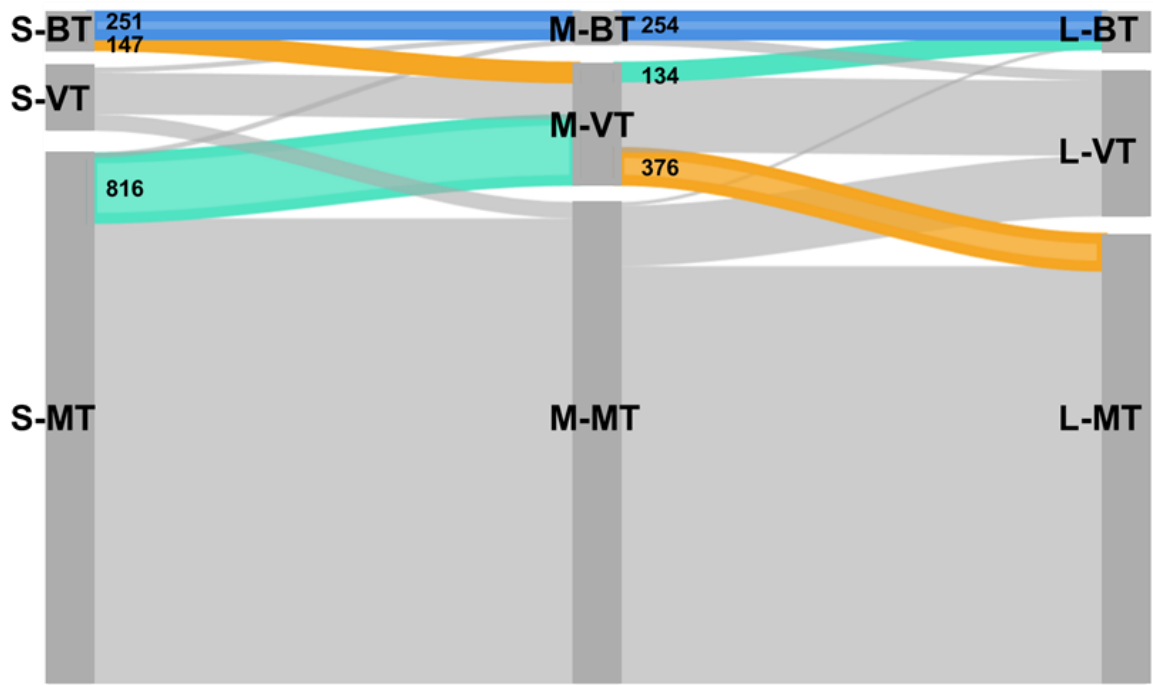

**Figure 3.** Time-variant impact of patented technologies

**Note:** The numbers inside the lines in each color represent the counts of patents corresponding to that color

## 4.2. RQ2: Does the impact of different time periods influence each other?

In RQ2, we examined whether shared parametric knowledge enhances the understanding of the time-variant technology impact. We developed MTL models using Keras for model development and optimization. Our MTL model comprises of shared and task-specific layers. The shared representation was learned through two fully connected layers with 128 and 64 units. Each task then branches into its own layers: Task 1 (short-term forecasting) employs two layers with 64 and 32 units; Task 2 (mid-term forecasting) has a single 64-unit layer; and Task 3 (long-term forecasting) uses two layers with 64 and 32 units. A total of 8,454 patents registered between 2006 and 2011 were utilized as the training set, whereas 2,397 patents registered in 2012 were reserved as the test set.

We explored the optimal architecture and hyperparameters with an exhaustive grid search using categorical cross-entropy loss as the loss function, SoftMax as the activation function, and a learning rate of 1e-3. Early stopping was implemented based on the aggregate validation loss across all the tasks. Batch training with a size of 32 and dropout rate of 0.5 was applied to the shared layers. Finally, the validity of the MTL models was investigated based on their ability to classify moderate, valuable, and breakthrough technologies, as shown in Tables 4 and A. 2. We evaluated the performance of the MTL models using stratified 5-fold cross-validation, with the sum of MCC scores across tasks serving as the performance evaluation metric to identify the best-performing model.

**Table 4.** Performance evaluation of MTL models for multiple time horizons.

| **(a) Short-term forecasting** | | | | |
|---|---|---|---|---|
| | **MT** | **VT** | **BT** | **Overall** |
| Accuracy | 0.9032 | 0.9366 | 0.9616 | 0.9007 |
| Precision | 0.9112 | 0.3125 | 0.5128 | 0.5788 |
| Recall | 0.9889 | 0.0342 | 0.215 | 0.4127 |
| F1 score | 0.9484 | 0.0617 | 0.303 | 0.4377 |
| MCC | 0.2373 | 0.0862 | 0.3157 | 0.2199 |
| DOR | 13.252 | 7.2211 | 32.9488 | 17.8073 |

**(b) Mid-term forecasting**

|           | MT     | VT     | BT      | Overall |
|-----------|--------|--------|---------|---------|
| Accuracy  | 0.864  | 0.8782 | 0.9683  | 0.8552  |
| Precision | 0.8844 | 0.2532 | 0.4186  | 0.5187  |
| Recall    | 0.9696 | 0.0791 | 0.2609  | 0.4365  |
| F1 score  | 0.9251 | 0.1205 | 0.3214  | 0.4557  |
| MCC       | 0.2372 | 0.0887 | 0.3151  | 0.194   |
| DOR       | 7.1645 | 3.0334 | 32.5129 | 14.2369 |

**(c) Long-term forecasting**

|           | MT     | VT     | BT      | Overall |
|-----------|--------|--------|---------|---------|
| Accuracy  | 0.8431 | 0.8469 | 0.9629  | 0.8265  |
| Precision | 0.8671 | 0.3237 | 0.2895  | 0.4934  |
| Recall    | 0.9596 | 0.1415 | 0.1507  | 0.4173  |
| F1 score  | 0.911  | 0.1969 | 0.1982  | 0.4354  |
| MCC       | 0.2898 | 0.1398 | 0.1913  | 0.2188  |
| DOR       | 7.7338 | 3.4808 | 15.0938 | 8.7695  |

We examined the effect of sharing parametric knowledge on estimating the dynamic technology impact by comparing the performance of MTL models with that of single-task learning (STL) models (Table 5). The hyperparameters of each STL model were optimized using a grid search. Overall, the STL models exhibited lower performance than the MTL model across most metrics. In particular, it is noteworthy that the ability to identify valuable or breakthrough technologies in the mid-term is enhanced in MTL models compared to STL models. For moderate technologies in mid-term forecasting, the STL model achieved a precision of 0.8769 and an MCC of 0.2087, which decreased by 0.0075 and 0.0285, respectively, compared with the MTL model. For breakthrough technologies, the STL model showed a recall of 0.1449, F1-score of 0.2, and MCC of 0.2011. Compared to the MTL model, which achieved 0.2609, 0.3214, and 0.3151, the STL model scores were lower by 0.1160, 0.1214, and 0.1140, respectively.

Table 5. Performance of STL models in mid-term forecasting.

|  | MT | VT | BT | Overall |
| --- | --- | --- | --- | --- |
| Accuracy | 0.8658 (+0.0058) | 0.8899 (+0.0117) | 0.9666 (-0.0017) | 0.8632 (-0.0080) |
| Precision | 0.8769 (-0.0075) | 0.2963 (+0.0431) | 0.3226 (-0.0960) | 0.4986 (-0.0201) |
| Recall | 0.9884 (+0.0188) | 0.0316 (-0.0475) | 0.1449 (-0.1160) | 0.3883 (-0.0482) |
| F1 score | 0.9293 (+0.0042) | 0.0571 (-0.0634) | 0.2000 (-0.1214) | 0.3955 (-0.0602) |
| MCC | 0.2087 (-0.0285) | 0.0663 (-0.0224) | 0.2011 (-0.1140) | 0.1585 (-0.0355) |
| DOR | 10.0888 (+2.9243) | 3.6520 (+0.6186) | 18.6198 (-13.8931) | 10.0888 (-4.1481) |

**Note:** Parentheses show the difference relative to MTL model performance

Similar trends were observed in the short- and long-term forecasting (Table A. 3). The improved performance in identifying breakthrough and valuable technologies suggests that MTL helps mitigate the bias of STL models, which tend to be biased towards moderate technology predictions (Table A. 4). For short- and long-term forecasting, the STL model showed a higher performance in identifying moderate technologies. The overall MCC of the MTL model was higher than that of the STL model across all time horizons. The largest performance gap between the two models is observed in the mid-term time horizon. In short-term forecasting, the small difference in citation counts between classes may make it challenging for both STL and MTL models to learn effectively. In the case of long-term forecasting, a sufficient difference in citation counts between classes could explain the smaller performance gap between the STL and MTL models.

**4.3. RQ3: Why does the impact of technology change over time?**

To answer RQ3, we investigated whether variables with a significant impact on the estimation of the dynamic technology impact change over time using SHAP analysis. Specifically, we employed Deep SHAP (Chen et al., 2021; Scott and Su-In, 2017), considering the modular structure of our models and computational complexity. For each time horizon, we identified the most important features with high global SHAP values for the breakthrough technologies (Figure 4).

     For short-term forecasting, scientific knowledge, technological recombination, and dependency on homogeneous technologies exhibit high SHAP values. For mid-term forecasting, the growth speed, technological breadth, and technology recombination showed high SHAP values for breakthrough technologies. For long-term forecasting, technological breadth, scientific knowledge, and the international priority range exhibited high SHAP values. The high impact of the international priority range suggests that having priority patents distributed across various countries is closely related to the validity of the patent, which in turn is associated with its long-term impact. In the mid-term, growth speed shows a notable influence, and referencing foundational technologies that have been in use for a long time tends to be related to higher technology impact. Summary plots for each task's BT prediction are presented in Figures A. 1, A. 2, and A. 3.

Although technology recombination and technological breadth exhibited high SHAP values in the short- and long-term predictions, a relationship between the magnitude of these variables and BT predictions was not clearly observed. In contrast, higher scientific knowledge had a stronger influence on BT predictions for both the long- and short-term. For long-term BT predictions, higher prior knowledge had a positive influence. For mid-term predictions, a broader international priority range had a lower influence on BT predictions. Similar to other time horizons, higher scientific knowledge had a positive influence. In addition, while technology breadth had high SHAP values, its relationship with BT predictions was not clearly identified.

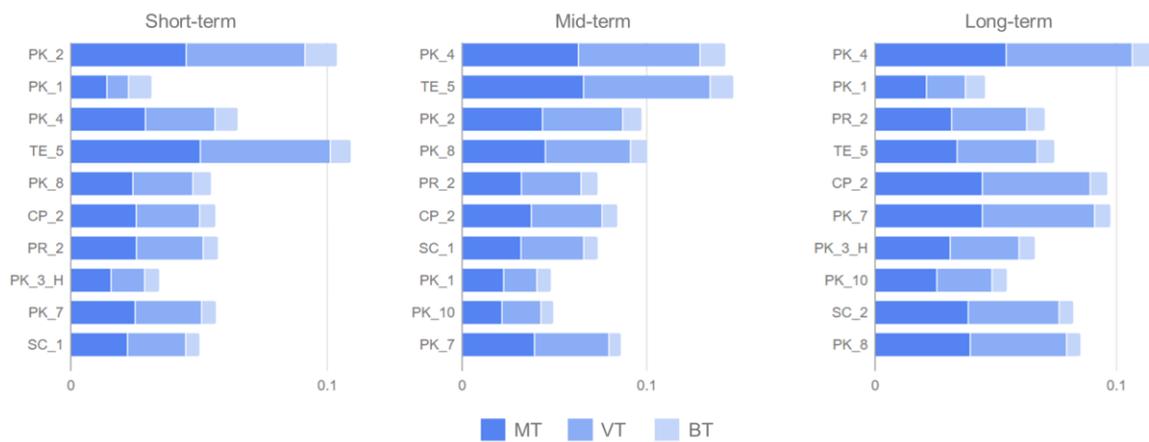

**Figure 4.** Important input variables for each time horizon

**Note:** Top 10 indicators ranked by SHAP value for BT prediction

We conducted an in-depth analysis for the sustained-impact patents group using SHAP analysis (Figure 5), while other patent groups are provided in Figures A. 4-5. For long-term BT prediction, scientific knowledge ranked highly and exhibited a strong positive correlation with the technology impact. Similarly, patents with high prior knowledge tend to be predicted as breakthrough technologies in the long-term. It suggests that a strong scientific basis and sufficient citation of prior knowledge are associated with a high technological impact. High technological breadth also showed a positive relationship with breakthrough patent predictions, indicating that referencing information from various technological fields is related to the high impact of technology. In contrast, dependency on homogeneous technologies has a negative influence. This implies that a heavy reliance on closely related prior technologies has a poor relationship with achieving a long-term high impact.

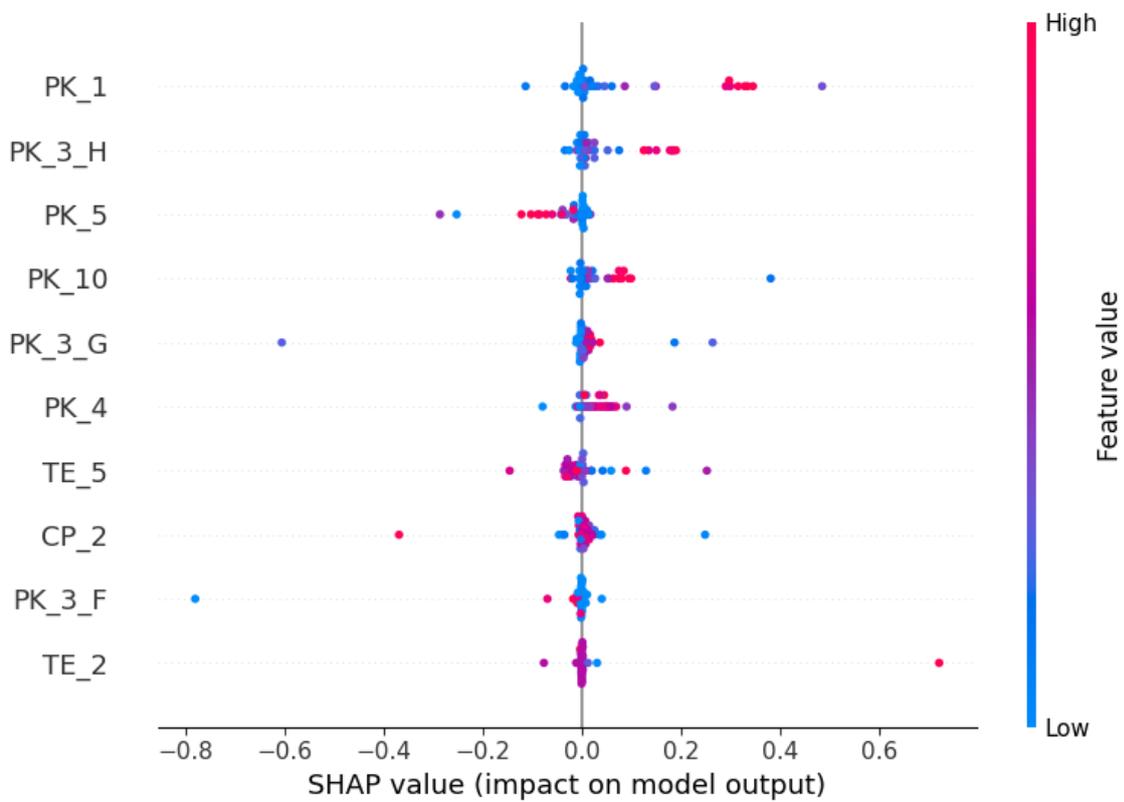

**Figure 5.** Summary plot for sustained-impact patents in long-term BT prediction.

# 5. Discussion

## 5.1. Validation of the proposed approach

To statistically validate our predictions, we examined whether the breakthrough technologies we identified exhibited higher post-hoc technology value metrics than those with lower impact. Three patent quality indicators, patent maintenance period, technology transfer count, and patent family size, were selected and extracted (Appendix A4). For each forecasting horizon, we investigated whether our predicted result–the technology impact–is correlated with other patent quality indicators. Jonckheere–Terpstra test was adopted to examine whether the rankings of the median quality indicators among the three technology impact classes were statistically significant (Table 6). A statistically significant trend was observed in the rankings of patent quality indicators across all time horizons. The technology impact classes demonstrated statistically significant ordinal trends for the three key indicators over the three horizons. Notably, technology transfer count and patent family size exhibited highly significant trends in all periods (p-value = 0.0000), while patent lifetime also showed significant trends with p-values of 0.0243 (short-term), 0.0169 (mid-term), and 0.0022 (long-term).

**Table 6.** Results of Jonckheere–Terpstra test.

| Horizon | Indicator | p-value |
|---|---|---|
| Short-term | Patent maintenance period | 0.0243 |
|  | Technology transfer count | 0.0087 |

| | Patent family size | 0.0000 |
|---|---|---|
| Mid-term | Patent maintenance period | 0.0169 |
| | Technology transfer count | 0.0003 |
| | Patent family size | 0.0000 |
| Long-term | Patent maintenance period | 0.0022 |
| | Technology transfer count | 0.0000 |
| | Patent family size | 0.0000 |

**Note:** This test is a nonparametric statistical method designed to evaluate whether there is a consistent trend, either increasing or decreasing, in the medians across three or more independent groups with ordinal data.

## 5.2. Implementation of the proposed approach

The proposed approach provides analysis results at the individual patent level; however, when combined with natural language processing techniques, it enables the derivation of interpretable technology impact scores across specific technical fields. We applied topic modeling techniques, including BERTopic, to analyze patent abstract texts, thereby identifying 22 distinct subdomains in battery technologies (Table A. 5). To assess the future impact of each topic, we analyzed the annual technology impact scores of the patents (Table 7). Excluding 2008, when 'direct methanol fuel cells' had the highest impact, 'energy storage devices' consistently demonstrated the highest average technological impact. Technology fields, such as 'fuel cell bipolar plates,' 'fuel cell stack design,' and 'fuel cell sealing components', consistently show low technological impact. In contrast, technological fields such as 'electrochemical cell components,' 'fuel cell catalysts,' and 'alkaline storage batteries' show growth trends.

Table 7. Annual technology impact scores for each topic.

| Topic | Year | | | | | | |
|---|---|---|---|---|---|---|---|
| | 2006 | 2007 | 2008 | 2009 | 2010 | 2011 | 2012 |
| battery pack design | 1.2813 | 1.6889 | 1.3922 | 1.4982 | 1.2935 | 1.4869 | 1.2312 |
| fuel cell power generation | 1.3490 | 1.2880 | 1.1746 | 1.1067 | 1.2903 | 1.1296 | 1.1145 |
| polymer electrolyte membranes | 1.1053 | 1.0588 | 1.3944 | 1.1263 | 1.1053 | 1.0672 | 1.1074 |
| electrochemical cell components | 1.3421 | 1.2353 | 1.2791 | 1.3269 | 1.3077 | 1.2917 | 1.6505 |
| cathode active materials | 1.0000 | 1.3200 | 1.3939 | 1.0851 | 1.1127 | 1.2581 | 1.3623 |
| lithium-ion battery electrodes | 1.0000 | 1.3214 | 1.2222 | 1.0000 | 1.0000 | 1.0769 | 1.1765 |
| solid oxide fuel cells | 1.1905 | 1.0000 | 1.0000 | 1.1143 | 1.0784 | 1.6842 | 1.3778 |
| fuel cell catalyst | 1.1667 | 1.6000 | 1.6667 | 1.4800 | 1.6190 | 1.2131 | 1.7451 |
| non-aqueous lithium electrolytes | 1.0000 | 1.0000 | 1.2353 | 1.0000 | 1.2889 | 1.0000 | 1.4815 |
| alkaline storage batteries | 1.0000 | 1.1905 | 1.0000 | 1.0000 | 1.0000 | 1.3077 | 1.4483 |
| fuel cell stack design | 1.0000 | 1.0000 | 1.0000 | 1.0000 | 1.2708 | 1.0000 | 1.0930 |
| hybrid electrolytes | 1.0000 | 1.0000 | 1.0000 | 1.4500 | 1.0000 | 1.1481 | 1.1081 |
| electrolyte electrode assembly | 1.1905 | 1.0000 | 1.0000 | 1.1538 | 1.0000 | 1.0000 | 1.1143 |
| fuel cell stack design | 1.0000 | 1.0000 | 1.0000 | 1.0000 | 1.0000 | 1.0000 | 1.0000 |
| anode active materials | 1.0000 | 1.0000 | 1.0000 | 1.0000 | 1.0000 | 1.2353 | 1.1212 |
| solid electrolyte | 1.3636 | 1.6429 | 1.0000 | 1.2000 | 1.0000 | 1.2105 | 1.0000 |
| fuel cell bipolar plate | 1.0000 | 1.0000 | 1.0000 | 1.0000 | 1.0000 | 1.0000 | 1.0000 |
| battery separators | 1.0000 | 1.0000 | 1.0000 | 1.0000 | 1.0000 | 1.7647 | 1.0000 |
| lithium battery negative electrodes | 1.0000 | 1.0000 | 1.0000 | 1.0000 | 1.0000 | 1.2857 | 1.0000 |
| direct methanol fuel cells | 1.4444 | 1.8889 | **2.0000** | 1.0000 | 1.0000 | 1.0000 | 1.0000 |
| energy storage devices | **4.0000** | **3.0000** | 1.0000 | **2.1250** | 1.0000 | **2.0000** | **2.4400** |
| fuel cell sealing components | 1.0000 | 1.0000 | 1.0000 | 1.0000 | 1.0000 | 1.0000 | 1.0000 |

**Note:** A weighted average of patent counts was calculated by assigning different weights to each impact class: breakthrough technologies (10), valuable technologies (5), and moderate technologies (1). The accumulated score for each topic was then normalized by dividing it by the number of patents within that topic. A technology impact score of 1 means that there are neither breakthrough nor valuable technologies.

# 6. Conclusion

This study proposed an MTL approach to enhance the understanding of the time-variant nature of technology impacts and the interrelationships between impacts across different time periods. The central tenet of this research is that capturing the dynamics of technology impacts enhances the prediction of time-variant impacts. Through a case study in the field of battery technology, we confirm that impact ratings can vary over time. By applying MTL, we enhanced the performance by improving the understanding of the dynamics of the technology impacts. In addition, through SHAP analysis, we investigated whether input features have a significant impact on the dynamic technology impact, which changes over time.

This study contributes to both academia and industry in several ways. From an academic perspective, this study is a pioneering attempt to use MTL to consider the relationships between impacts across different time horizons at the early stage of patent registration. We demonstrated the potential of MTL in technology impact analysis and contributed to the precision of potential technology indicators by enabling predictions of future technology impact over time. From a practical perspective, the proposed method can assist in decision-making for intellectual property management. Forecasting the time-variant technology impact enables the development of detailed strategies, such as licensing agreements, technology transfer, technology valuation, and patent maintenance. In addition, this approach aids in identifying emerging technology areas and supporting R&D decisions by adjusting plans based on whether an emerging area has long-term or short-term impact potential.

Despite the effectiveness of the proposed approach, this study had several limitations. First, additional input features, such as textual information, can be incorporated to enhance the technological understanding of patented inventions (Hong et al., 2022). Moreover, network-based indicators that capture a patent's characteristics from a broader perspective can be incorporated in terms of the technological ecosystem (Choi et al., 2023). Second, long-term predictions are challenging for newly issued patents, as they are less than 10 years old, and thus have insufficient citation data. Because MTL allows the use of different datasets for each task, future research could explore long-term predictions using older patents and short-term predictions using recent patents. Finally, the proposed approach currently relies only on forward citation counts as a proxy for technological value. Future work could incorporate additional indicators, such as technology transaction data (Kim et al., 2022; Ko et al., 2019) and patent maintenance (Choi et al., 2020), within the MTL framework to enhance the multifaceted understanding of different technology value indicator.

# Appendix

## A1. Description for patent indicators

In this study, 46 indicators were selected based on literature. These indicators are categorized into six groups: scope and coverage, priority, completeness, development effort and capabilities, technological environment, and prior knowledge. Ex-ante indicators, which can be collected at the early stage of patent registration, were extracted through database queries.

### A1.1. Scope and coverage

To reflect the international scope of prior patents, the number of different nations in the patent's backward citation information was used. This is because the origins of the technologies that underpin an invention can potentially impact a patent's value. The scope of the core and peripheral technological components is calculated as the number of independent and dependent claims in the patent. Independent claims of a patent are the most fundamental knowledge of the patent, and dependent claims provide detailed information about cited claims (Reitzig, 2004; Tong and Frame, 1994). Therefore, the more claims a patent has, the more likely it is to be registered and impactful (Fischer and Leidinger, 2014; Lanjouw and Schankerman, 2004). Moreover, the type and number of claims are related to the costs of patent registration (Jeong et al., 2016; Lanjouw and Schankerman, 2001). Likewise, the specificity of independent claims is related to the scope of legal protection or novelty of the invention. The average specificity of the core technological components can be calculated as the average number of words contained in a patent's independent claims. The number of different IPCs assigned to a patent was calculated to reflect the technical coverage of the system. Technical coverage can assist in identifying patent value (Lee et al., 2009).

### A1.2. Priority

The priority of patents protects the initial filing date of an invention, validating its early stage innovation and novelty, thereby enhancing the value of the patent (Choi et al., 2020; Su et al., 2011). Therefore, the priority date is crucial for assessing the overall worth of a patent. In this regard, the priority intensity and its international range are reflected. These were calculated as the number of priority patents for patents and different nations.

**A1.3. Development effort and capabilities**

Six indicators were employed as inputs for the development effort and capabilities. Because patent maintenance is affected by the capabilities of assignees (Lai and Che, 2009), the number of assignees is calculated to reflect the size of contributors. Likewise, the number of non-US assignees is used to quantify the contributions of foreign applicants. To account for the varying tendencies in patent maintenance or value across nations, the number of different nations of assignees is utilized as a proxy for the international distribution of contributors. In addition, given that inventions involving multiple inventors are often more valuable than those involving a single inventor (Ernst, 2003; Ma and Lee, 2008), the number of inventors, non-US inventors, and number of different nations of inventors are utilized to reflect the efforts of inventors or foreign inventors and the degree of international cooperation.

**A1.4. Completeness**

Two indicators are used to consider the completeness of inventions: grant time lag and specificity of the technical summary. The interval between a patent's filing and issuance dates, representing the patent examination period, can be used as an indicator of grant time lag. A longer examination period may reflect significant stakeholder efforts to strengthen the patent, potentially suggesting a higher technological value or complexity (Hikkerova et al., 2014). The specificity of the technical summary is measured by the word count of the patent's abstract, which includes the core aspects of the invention. The level of detail can indicate the scope or value of the patent, with more specific descriptions often reflecting broader or more valuable claims.

**A1.5. Technology environment**

Five environmental indicators were used, as the technological environment may relate to patent maintenance and R&D strategies (Choi et al., 2023; Trappey et al., 2012). To account for the activity of technology fields, the average number of patents issued annually in the relevant technology fields was utilized. Similarly, the average number of cumulative patents issued in the relevant technology fields is used to quantify the size of the technology fields (Lai and Che, 2009). In addition, the average number of applicants issuing patents by issuance year in the relevant technology fields is utilized to reflect the competitiveness of the technology fields (Fabry et al., 2006). The frequency of IPC codes at the section level was used to reflect the technology fields in the system. The growth speed indicator is determined by measuring the median time gap between the filing dates of prior and current patents, referred to as the technology cycle time (Bierly and Chakrabarti, 1996; Kayal and Waters, 1999). This reflects the pace at which innovation progresses within the relevant technology field.

**A1.6. Prior knowledge**

For prior knowledge, ten measures were employed. The number of non-patent citations is employed to represent the scientific knowledge of an invention because a stronger scientific basis is often linked to more innovative and influential technologies (Cozzens et al., 2010; Rotolo et al., 2015; Trajtenberg, 1990a). To measure the prior knowledge of inventors and assignees, the average number of patents they have previously been involved in is used, reflecting their experience and its impact on patent value. Core and peripheral knowledge are assessed based on the assignee's prior activities (Harhoff et al., 1999), either within the same technological area (core) or other fields (peripheral) (Seol et al., 2023), representing the accumulated knowledge that reflects both technological and commercial focus (Chung et al., 2021; Meyer, 2006). Many studies have proposed that the novelty of a technology arises from the recombination and synthesis of existing technologies (Fleming and Sorenson, 2001; Lee and Lee, 2019). Owing to the complexity of calculating all IPC combinations for all patents in the relevant IPC codes (Kwon and Geum, 2020), the technology recombination of the invention is measured as the ratio of the IPC codes, which is the recombination of the IPC from prior patents. The number of technological classes a patent cites can be used as a proxy measure of technological value (Kwon and Geum, 2020; Shane, 2001; Verhoeven et al., 2016). Technological breadth refers to the diversity of IPC codes associated with patents and prior patents. This reflects how broad the technological scope of a patent is (Choi and Yoon, 2022). In this study, technological breadth was modified and calculated as the number of technological classes of patent sites. Dependency on homogeneous technologies is calculated by counting the number of patents cited within the same battery technology field, reflecting the extent to which the invention relies on closely related prior technologies. Prior knowledge was represented by the number of backward citations.

### A2. Description for performance metrics

Initially, class-specific accuracy and overall average accuracy are computed (Equations 6-7).

$$Accuracy_i = \frac{TP_i + TN_i}{TP_i + TN_i + FP_i + FN_i} \tag{6}$$

$$\text{Average accuracy} = \frac{\sum_{i=1}^{l}((TP_i + TN_i)/(TP_i + TN_i + FP_i + FN_i))}{l} \tag{7}$$

In this context, the terms true positive ($TP_i$), true negative ($TN_i$), false positive ($FP_i$), and false negative ($FN_i$) for a given class $i$ represent correctly predicted positives, correctly predicted negatives, negatives wrongly predicted as positives, and positives wrongly predicted as negatives, respectively. The total number of classes is denoted by $l$. In a multi-class classification problem, accuracy measures the proportion of correctly classified patents. However, it can be misleading in imbalanced datasets, where certain grades dominate, and high accuracy may mask poor performance on less frequent classes. Therefore, although accuracy is a useful overall metric, it should be complemented with precision, recall, and F1-score for a complete assessment (Equations 8-10).

$$Precision_i = \frac{TP_i}{TP_i+FP_i} \tag{8}$$

$$Recall_i = \frac{TP_i}{TP_i+FN_i} \tag{9}$$

$$F1-score_i = \frac{2TP_i}{2TP_i+FP_i+FN_i} \tag{10}$$

### A3. Description for citation patterns

The pattern of sustained-impact patents is attributed to (1) enduring technological relevance, where the patents continue to effectively address fundamental needs or challenges; (2) broad applicability, allowing the technology to be widely used across various applications and industries; (3) persistent market demand, where ongoing interest and adoption maintain the significance of technology; and (4) continuous innovation, in which technology evolves and adapts to new developments, preserving its impact and value over time. The reasons for peak-and-fade patents, in impact over time, include (1) technological obsolescence, where advancements in newer technologies render the initial innovations less relevant; (2) market saturation, where widespread adoption of the technology reduces its incremental significance; (3) increased competition, as alternative solutions emerge and overshadow the original technology; and (4) shifts in industry focus, where changes in priorities or consumer preferences lead to diminished relevance of the technology (Lee et al., 2017). The reasons for late-blooming patents include (1) high technological uncertainty, such as discrepancies between computational predictions and experimental performance; (2) limited practical applicability, including challenges in scaling from laboratory experiments to industrial production; (3) low initial market acceptance, exemplified by fluctuations in the electric vehicle industry; and (4) high development costs, characterized by the use of high-performance materials with substantial price tags (Haupt et al., 2007).

**A4. Description for technology value indicators**

First, the patent maintenance period reflects the period during which the rights to a given technology are actively maintained. Maintaining a patent for an extended period suggests that the holder perceives sustained business or industrial value in the technology (Choi et al., 2020). Second, technology transfer, as evidenced by the transaction history of patent rights, is a key indicator of a patent's commercial value and market potential (Kim et al., 2022). Finally, the patent family—a collection of patents filed across different jurisdictions to protect the same invention—provides insight into an applicant's efforts to safeguard their innovation in multiple markets.

# Tables

**Table A. 1.** Distribution of patents by number of citations.

**(a) The number of patents based on the number of citations received within three years.**

| Number of citations | Number of patents | Percentage of patents | Cumulative percentage |
|---|---|---|---|
| 0 | 7635 | 70.36% | 70.36% |
| 1 | 1969 | 18.15% | 88.51% |
| 2-3 | 849 | 7.82% | 96.33% |
| 4-7 | 307 | 2.83% | 99.16% |
| 8-61 | 91 | 0.84% | 100% |

**(b) The number of patents based on the number of citations received within five years.**

| Number of citations | Number of patents | Percentage of patents | Cumulative percentage |
|---|---|---|---|
| 0 | 5481 | 50.51% | 50.51% |
| 1-2 | 3456 | 31.85% | 82.36% |
| 3-8 | 1522 | 14.03% | 96.39% |
| 9-17 | 289 | 2.66% | 99.05% |
| 18-147 | 103 | 0.95% | 100% |

**(c) The number of patents based on the number of citations received within ten years.**

| Number of citations | Number of patents | Percentage of patents | Cumulative percentage |
|---|---|---|---|
| 0 | 3300 | 30.41% | 30.41% |
| 1-5 | 5152 | 47.48% | 77.89% |
| 6-23 | 1964 | 18.10% | 95.99% |
| 24-53 | 326 | 3.00% | 99.0% |
| 54-880 | 109 | 1.00% | 100% |

Table A. 2. Confusion matrix of MTL models for multiple time horizons.

(a) Short-term forecasting

| | Predicted as MT | Predicted as VT | Predicted as BT |
|---|---|---|---|
| **MT** | 2134 | 7 | 17 |
| **VT** | 129 | 5 | 2 |
| **BT** | 69 | 4 | 20 |

(b) Mid-term forecasting

| | Predicted as MT | Predicted as VT | Predicted as BT |
|---|---|---|---|
| **MT** | 2012 | 49 | 14 |
| **VT** | 222 | 20 | 11 |
| **BT** | 41 | 10 | 18 |

(c) Long-term forecasting

| | Predicted as MT | Predicted as VT | Predicted as BT |
|---|---|---|---|
| **MT** | 1925 | 68 | 13 |
| **VT** | 259 | 45 | 14 |
| **BT** | 36 | 26 | 11 |

Table A. 3. Performance of STL models in short-term and long-term forecasting (Parentheses show the difference relative to MTL model performance).

(a) Short-term forecasting

|  | MT | VT | BT | Overall |
|---|---|---|---|---|
| Accuracy | 0.9045 (+0.0013) | 0.9395 (+0.0029) | 0.9641 (+0.0025) | 0.9040 (+0.0033) |
| Precision | 0.9170 (-0.0058) | 1.0 (+0.6875) | 0.7059 (+0.1931) | 0.8704 (+0.2916) |
| Recall | 0.9981 (+0.0092) | 0.0068 (-0.0274) | 0.1290 (-0.0860) | 0.3780 (-0.0347) |
| F1 score | 0.9495 (+0.0011) | 0.0136 (-0.0481) | 0.2182 (-0.0848) | 0.3938 (-0.0439) |
| MCC | 0.1969 (-0.0404) | 0.0802 (-0.0060) | 0.2919 (-0.0238) | 0.1949 (-0.0250) |
| DOR | 33.5066 (+20.2546) | ∞ (>7.2211) 7.2211 (-∞) | 68.1185 (+35.1697) | ∞ (>17.8073) 17.8073 (-∞) |

**(b) Long-term forecasting**

|  | MT | VT | BT | Overall |
|---|---|---|---|---|
| Accuracy | 0.8461 (+0.003) | 0.8594 (+0.0125) | 0.9721 (+0.0046) | 0.8275 (+0.010) |
| Precision | 0.8533 (-0.0138) | 0.3227 (-0.010) | 0.5381 (+0.1243) | 0.5269 (+0.0335) |
| Recall | 0.9855 (+0.0159) | 0.0503 (-0.0912) | 0.1644 (+0.0137) | 0.4001 (-0.0172) |
| F1 score | 0.9146 (+0.0036) | 0.0867 (-0.1102) | 0.2015 (+0.0371) | 0.4122 (-0.0232) |
| MCC | 0.2386 (-0.0512) | 0.0787 (-0.0611) | 0.2469 (+0.0556) | 0.1799 (-0.0389) |
| DOR | 10.2259 (+2.4921) | 3.0940 (-0.3868) | 26.6962 (+11.6024) | 13.3387 (+4.5692) |

Table A. 4. Confusion matrix of STL in short-term and long-term forecasting.

**(a) Short-term forecasting**

|  | Predicted as MT | Predicted as VT | Predicted as BT |
|---|---|---|---|
| **MT** | 2154 | 0 | 4 |
| **VT** | 144 | 1 | 1 |
| **BT** | 81 | 0 | 12 |

**(b) Mid-term forecasting**

|  | Predicted as MT | Predicted as VT | Predicted as BT |
|---|---|---|---|
| **MT** | 2051 | 13 | 11 |
| **VT** | 235 | 8 | 10 |
| BT | 53 | 6 | 10 |
| **(c) Long-term forecasting** | | | |
| | **Predicted as MT** | **Predicted as VT** | **Predicted as BT** |
| **MT** | 1977 | 23 | 6 |
| **VT** | 291 | 16 | 11 |
| **BT** | 49 | 12 | 12 |

**Table A. 5.** Results of BERTopic modeling

| Topic | Count | Main keywords |
|---|---|---|
| battery pack design | 2172 | 'battery pack', 'secondary battery', 'battery module', 'battery cell', 'power supply', 'battery case', 'electronic device', 'battery cells', 'rechargeable battery', 'electrode assembly' |
| fuel cell power generation | 1968 | 'fuel cell', 'cell stack', 'fuel cell stack', 'fuel gas', 'power generation', 'fuel cell includes', 'includes fuel', 'cell fuel', 'hydrogen gas', 'fuel cell fuel' |
| polymer electrolyte membranes | 730 | 'polymer electrolyte', 'electrolyte membrane', 'polymer electrolyte membrane', 'membrane electrode', 'membrane electrode assembly', 'catalyst layer', 'electrode assembly', 'solid polymer', 'solid polymer electrolyte', 'polymer electrolyte fuel' |
| electrochemical cell components | 420 | 'electrochemical cell', 'electrochemical device', 'second electrode', 'active material', 'electrode active', 'electrode active material', 'electrode plate', 'electrochemical cells', 'electrode strip', 'material layer' |
| cathode active materials | 336 | 'active material', 'cathode active', 'composite oxide', 'transition metal', 'cathode active material', 'positive electrode', 'positive electrode active', 'electrode active', 'electrode active material', 'material lithium' |
| lithium-ion battery electrodes | 259 | 'negative electrode', 'positive electrode', 'lithium secondary', 'secondary battery', 'lithium ion', 'lithium secondary battery', 'electrode assembly', 'lithium ion secondary', 'ion secondary', 'electrode plate' |
| solid oxide fuel cells | 237 | 'solid oxide', 'oxide fuel', 'solid oxide fuel', 'oxide fuel cell', 'fuel cell', 'electrode layer', 'electrolyte layer', 'fuel electrode', 'oxide fuel cells', 'air electrode' |
| fuel cell catalyst | 235 | 'supported catalyst', 'catalyst fuel', 'electrode catalyst', 'catalyst fuel cell', 'catalyst particles', 'metal catalyst', 'noble metal', 'fuel cell', 'method preparing', 'fuel cells'] |
| non-aqueous lithium electrolytes | 193 | 'non aqueous', 'organic solvent', 'lithium salt', 'lithium secondary', 'non aqueous electrolyte', 'aqueous electrolyte', 'lithium secondary battery', 'secondary battery', 'electrolyte solution', 'electrolyte lithium' |
| alkaline storage batteries | 183 | 'alkaline electrolyte', 'nickel hydroxide', 'manganese dioxide', 'positive electrode', 'alkaline battery', 'negative electrode', 'alkaline storage', 'alkaline storage battery', 'zinc air', 'storage battery' |
| fuel cell stack design | 182 | 'fuel cell stack', 'cell stack', 'fuel cell', 'fuel cells', 'end plate', 'stack includes', 'cell stack includes', 'cell assembly', 'plurality fuel', 'end plates' |
| hybrid electrolytes | 165 | 'non aqueous', 'non aqueous electrolyte', 'aqueous electrolyte', 'negative electrode', 'positive electrode', 'nonaqueous electrolyte', 'electrolyte secondary', 'electrolyte secondary battery', 'active material', 'electrode active' |
| electrolyte electrode assembly | 157 | 'fuel cell', 'electrolyte electrode', 'fuel gas', 'fuel cell includes', 'electrolyte electrode assembly', 'electrode assembly', 'catalyst layer', 'oxygen containing gas', 'oxygen containing', 'containing gas' |
| fuel cell stack design | 138 | 'separator fuel', 'fuel cell separator', 'cell separator', 'separator fuel cell', 'fuel cell', 'metal separator', 'fuel gas', 'stainless steel', 'contact resistance', 'separator member' |
| anode active materials | 120 | 'anode active', 'anode active material', 'active material', 'active material layer', 'material layer', 'anode current collector', 'charge discharge', 'anode current', 'provided anode', 'collector anode' |
| solid electrolyte | 111 | 'solid electrolyte', 'polymer electrolyte', 'electrochemical devices', 'solid polymer', 'electrolyte composition', 'gel electrolyte', 'ionic liquid', 'solid polymer electrolyte', 'high ion', 'compound having' |
| fuel cell bipolar plate | 98 | 'bipolar plate', 'plate fuel', 'bipolar plates', 'plate fuel cell', 'flow field', 'fuel cell', 'plate includes', 'surface layer', 'bipolar plate includes', 'flow channels' |
| battery separators | 92 | 'battery separator', 'porous substrate', 'inorganic particles', 'separator layer', 'binder polymer', 'porous film', 'nonwoven fabric', 'separator electrode', 'pores porous', 'separator battery' |
| lithium battery negative electrodes | 89 | 'negative electrode', 'active material', 'material layer', 'active material layer', 'negative electrode active', 'secondary battery', 'negative electrode material', 'collector active material', 'collector active', 'negative electrode lithium' |
| direct methanol fuel cells | 83 | 'methanol fuel', 'direct methanol', 'methanol fuel cell', 'direct methanol fuel', 'liquid fuel', 'fuel cell', 'fuel cartridge', 'methanol solution', 'fuel reservoir', 'catalyst layer' |
| energy storage devices | 64 | 'energy storage', 'storage device', 'energy storage device', 'collector foil', 'second electrodes', 'storage device includes', 'device includes', 'second material', 'electrode electrical', 'second electrode' |
| fuel cell sealing components | 63 | 'seal member', 'fuel cell', 'core portion', 'seal material', 'seal portion', 'bipolar plate', 'flow field', 'sealing surface', 'body 11', 'sealing structure' |
| fuel cell electrode materials | 2756 | 'fuel cell', 'negative electrode', 'active material', 'positive electrode', 'secondary battery', 'electrode active', 'electrode active material', 'fuel cells', 'membrane electrode', 'electrode assembly' |

**Figures**

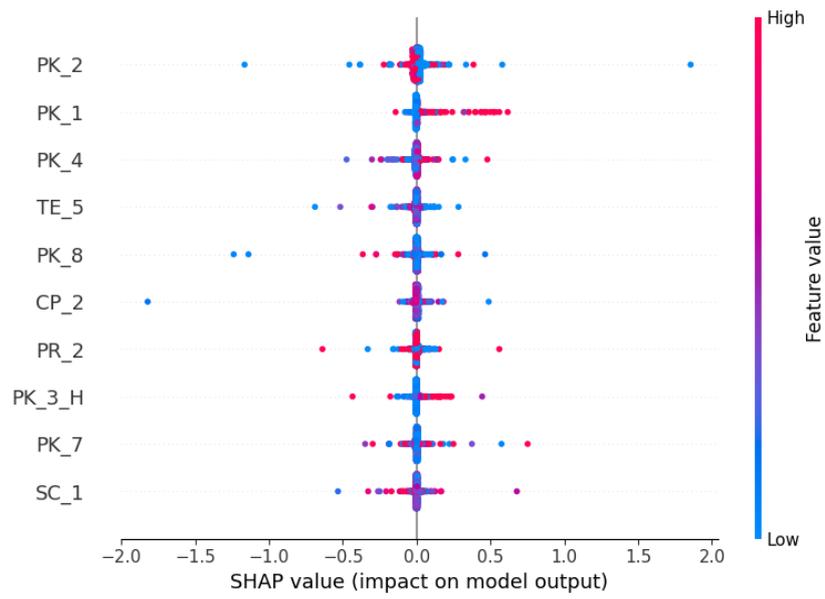

**Figure A. 1.** Summary plot for short-term BT forecasting.

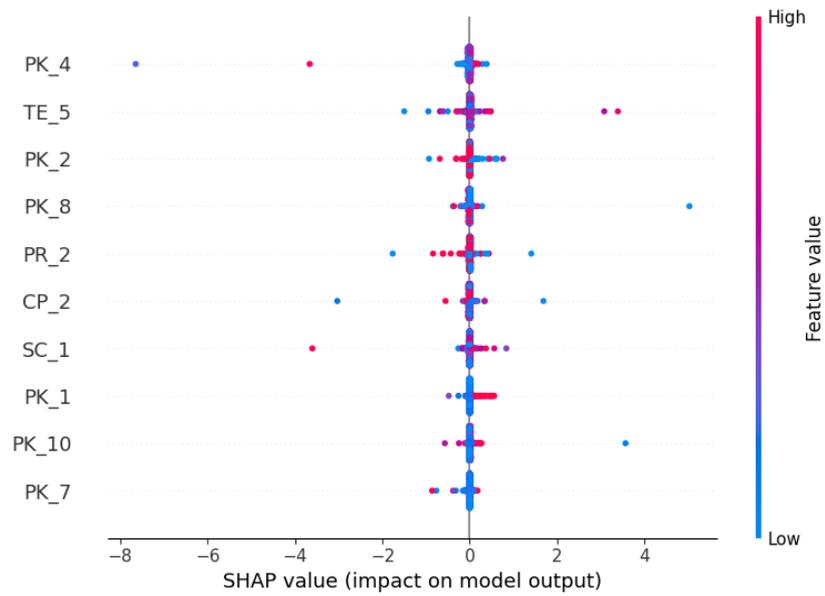

**Figure A. 2.** Summary plot for mid-term BT prediction.

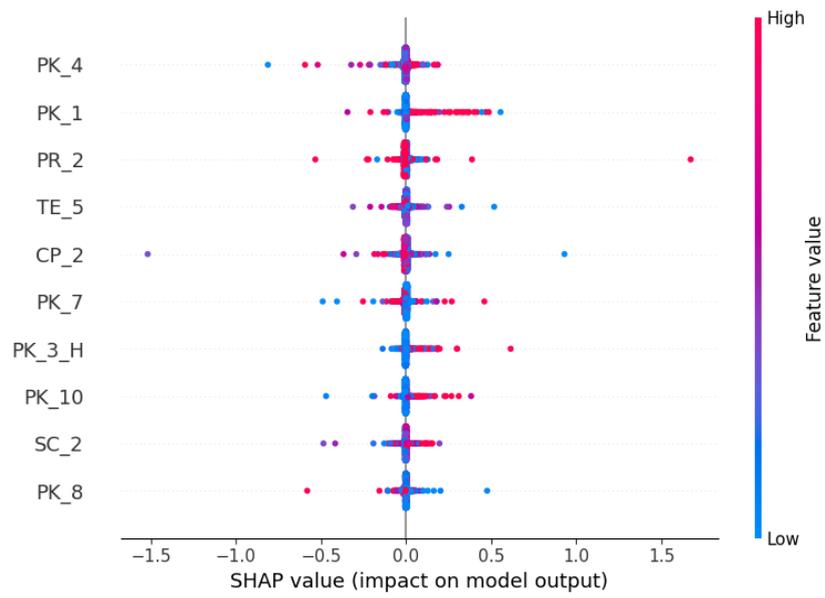

**Figure A. 3.** Summary plot for long-term BT prediction.

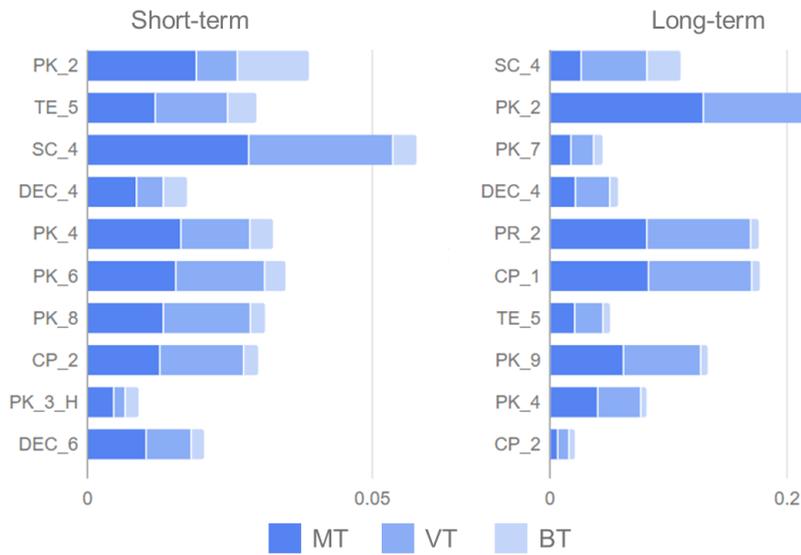

**Figure A. 4.** Important input variables for multiple time horizons of peak-and-fade patents.

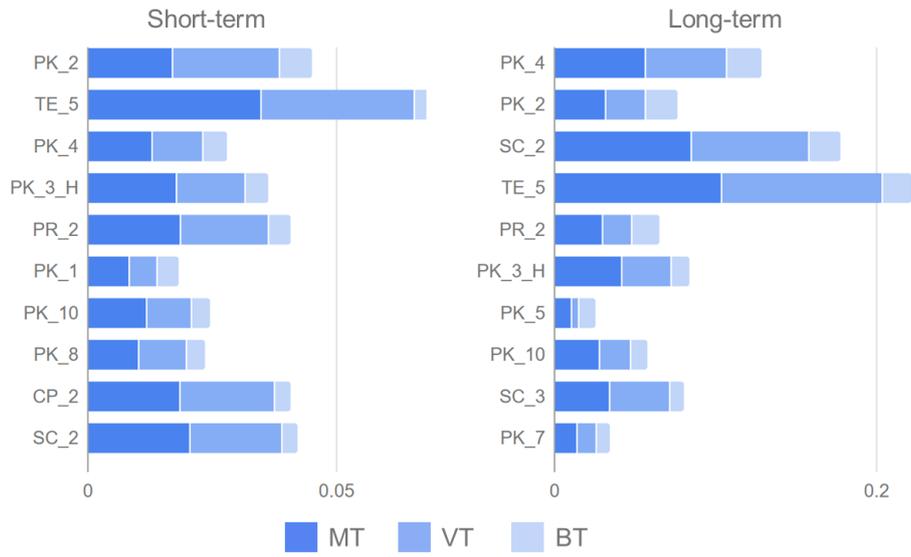

**Figure A. 5.** Important input variables for multiple time horizons of late-blooming patents.